\documentclass[letterpaper, 10 pt, conference]{ieeeconf}  

\IEEEoverridecommandlockouts
\usepackage{times}
\usepackage{graphicx}
\usepackage{booktabs}
\usepackage{amsmath,amssymb}
\usepackage{cite}
\usepackage[misc]{ifsym}
\usepackage[hidelinks]{hyperref}

\title{\LARGE \bf
LLM-Guided Future Hypotheses for Horizon-Aware Exploration in Multi-Step Robot Manipulation
}

\author{
Mohammad Khoshnazar$^{1}$, Andrew Melnik$^{1}$, Michael Beetz$^{1}$%
\thanks{$^{1}$Institute of Artificial Intelligence, University of Bremen, Germany.}
}

\begin{document}

\maketitle
\thispagestyle{empty}
\pagestyle{empty}

\begin{abstract}
Multi-step robot manipulation requires acting under uncertainty about how the scene will evolve, making exploration and policy adaptation challenging. We study whether short-horizon, task-consistent future videos can provide useful structured priors for control and reinforcement-learning fine-tuning.

We formalize this idea through \textbf{Future-Experience Conditioning (FEC)}, a simple interface that conditions closed-loop policies on a latent representation of a short future video. In our simulation setup, future clips are generated in three stages, an LLM reasoner operating over a task ontology initialized from the current scene state, a robot-free digital-twin rollout of the intended object motion, and a mask-free video diffusion model that synthesizes a robot-consistent future clip without requiring segmentation at inference.

We instantiate this future-conditioning interface primarily with \textbf{BC} and \textbf{BC+RL}, and compare against a future-conditioned Streaming Flow Policy (SFP)~\cite{jiang2025sfp} baseline on RoboCasa and CALVIN under \textbf{NoFuture}, \textbf{GTFuture}, \textbf{GenFuture}, and \textbf{WrongFuture}. Generated futures improve performance over no-future conditioning, while mismatched futures degrade it, and our \textbf{BC+RL} instantiation achieves the strongest overall results. An average BC+RL learning-curve analysis across 8 CALVIN tasks further shows that \textbf{GTFuture} improves fastest, \textbf{GenFuture} improves earlier and to a higher level than \textbf{NoFuture}, and \textbf{WrongFuture} remains at zero throughout training. These results suggest that short-horizon future videos can serve as useful structured priors for exploration and policy adaptation under imperfect future predictions. \url{https://enact2026.github.io/}
\end{abstract}

\begin{figure*}[t]
  \centering
  \includegraphics[width=\textwidth]{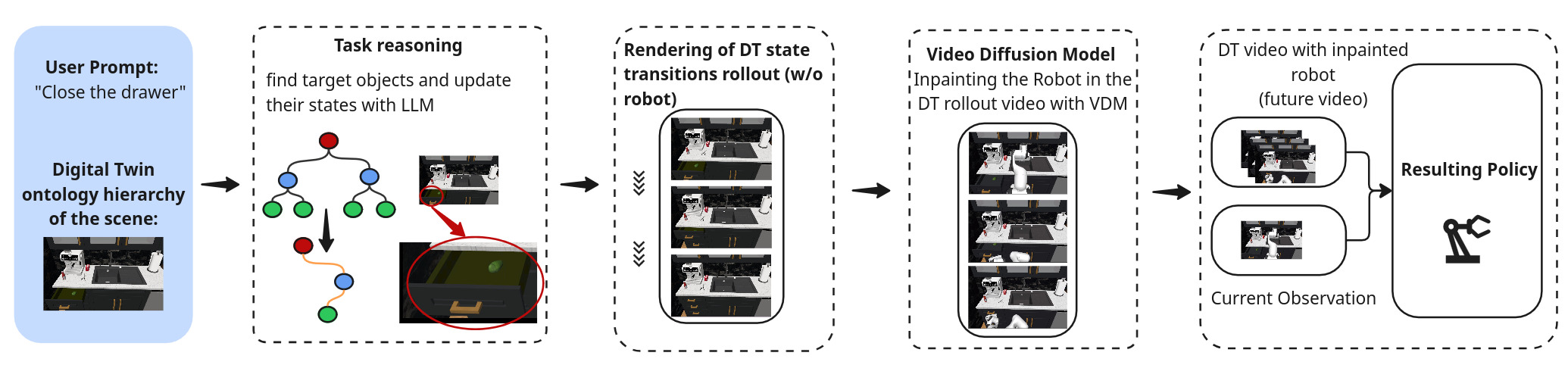}
  \caption{Overview of our \textbf{Future-Experience Conditioning (FEC)} pipeline. From the task prompt and current scene state, an ontology-guided LLM infers the target object, interaction part, desired state transition, and rollout constraints. This drives a robot-free digital-twin rollout (``transparent'' video). A mask-free video diffusion model then inpaints the robot to generate a task-consistent future clip. The policy conditions on future latents together with recent observations.}
  \label{fig:framework_overview}
\end{figure*}

\section{Introduction}
Many manipulation tasks require more than reactive control. To open a drawer, move a light switch, or push an object into a drawer, a robot must account for how present actions affect future interaction states over multiple steps. Success depends on delayed consequences such as contact timing, articulation response, and task progress. These challenges are broadly reflected in recent surveys of deep reinforcement learning and contact-rich robotic manipulation~\cite{han2023drlmanip,elguea2023contactrich}.

Recent generative video models make it possible to use a short visual hypothesis of how the scene and manipulated object should evolve. If informative, such a future can improve contact timing, disambiguate intent, and help recovery from drift. The challenge is that the future signal used during training differs from the generated frames available at test time and may be noisy or misaligned temporally. More broadly, this question sits at the intersection of visual foresight, latent world models, and large-scale robot policy learning~\cite{finn2017foresight,hafner2019planet,ha2018worldmodels,hafner2019dreamer,brohan2023rt1,brohan2023rt2,octo2024octo}.

In this work, we study short-horizon future hypotheses as a mechanism for horizon-aware exploration and policy adaptation in multi-step manipulation. We formalize this idea with \textbf{Future-Experience Conditioning (FEC)}, which conditions a closed-loop policy on a latent representation of a short-horizon future clip. Our framework takes a task command and current scene state, initializes a task ontology, and uses an \textbf{LLM reasoner} operating over this ontology to infer what object should be manipulated, which interaction part is relevant, and how the object state should change. It then produces a robot-free digital-twin rollout for the intended object motion and uses a mask-free video diffusion model to inpaint the robot into this rollout as a future video. This video is encoded into a future latent and provided, together with recent observations, to the controller for action generation (Fig.~\ref{fig:framework_overview}).

Generated future clips provide a form of \emph{horizon-aware exploration}, allowing the controller to condition on plausible near-future task evolution rather than acting blindly. By comparing correct and mismatched future hypotheses, we study when such structured priors help or hinder closed-loop manipulation.

This paper focuses on the role of future-conditioned visual priors in \emph{simulation}. Our goal is not to claim real-world transfer, end-to-end digital-twin construction from raw sensory input alone, or continuous online twin updates during closed-loop execution. Instead, we study whether generated short-horizon future videos can improve policy execution and reinforcement-learning-based adaptation under controlled train and test future mismatch.

Our main controller is \textbf{BC+RL}, while a future-conditioned \textbf{SFP} model is trained as a comparison baseline under the same future-conditioning interface.

\subsection{Contributions}
\begin{itemize}
  \item \textbf{Future hypotheses as structured priors:} We study short-horizon, task-consistent future videos as conditioning signals for multi-step robot manipulation and reinforcement-learning fine-tuning.
  \item \textbf{Future-Experience Conditioning (FEC):} We introduce a simple future-conditioning interface that compresses short-horizon future videos into temporally binned latent representations for closed-loop policy conditioning.
  \item \textbf{Empirical study with BC+RL as the main instantiation:} On RoboCasa and CALVIN, we evaluate BC, BC+RL, and a future-conditioned SFP comparison baseline under NoFuture, GTFuture, GenFuture, and WrongFuture. Our BC+RL instantiation is strongest overall, and an average BC+RL learning-curve analysis across 8 CALVIN tasks shows that informative generated futures can improve learning speed during RL fine-tuning while mismatched futures hinder adaptation.
\end{itemize}

\section{Related Work}
\label{sec:related}

Exploration is central to multi-step decision making in reinforcement learning and robotics, especially in manipulation where success depends on discovering interaction-relevant contacts and delayed consequences. Recent surveys review deep RL for robotic manipulation and contact-rich interaction settings, as well as broader reinforcement-learning applications across robotics domains~\cite{han2023drlmanip,elguea2023contactrich,tezerjani2024rlslam}. Our work is most closely related to approaches that guide behavior with structured predictions of future task evolution.

Visual anticipation methods predict future observations for planning and control~\cite{finn2017foresight}, while latent world models learn compact dynamics for longer-horizon behavior~\cite{hafner2019planet,ha2018worldmodels,hafner2019dreamer}. Sequence-modeling and generative control perspectives such as Decision Transformer and Diffuser are also closely related to the broader idea of exploiting predicted future structure for action selection~\cite{chen2021decision,janner2022diffuser}. More recent robotics approaches use generated futures either as intermediate visual goals~\cite{luo2025grounding} or in closed-loop replanning pipelines~\cite{bu2024clover}. Our setting differs in focusing on \emph{imperfect future supervision}, where the future signal used during training differs from the generated future available at test time.

For future generation, we build on diffusion-based video modeling and inpainting~\cite{ho2020ddpm,ho2022videodiffusion,lugmayr2022repaint}, specifically CogVideoX~\cite{yang2024cogvideox} and VideoPainter~\cite{bian2025videopainter}, to construct a mask-free VDM conditioned on a robot-free digital-twin rollout. Our work is also adjacent to recent vision-language and digital-twin style manipulation systems that use high-level reasoning or imagined goal states for rearrangement and control~\cite{kapelyukh2024dream2real,ahn2022saycan}.

For robot control, our main controller is a behavior-cloning pipeline followed by reinforcement-learning fine-tuning (BC+RL), together with a BC-only variant. We additionally adapt Streaming Flow Policy (SFP)~\cite{jiang2025sfp} to the same future-conditioning interface as a comparison baseline. More broadly, the paper is related to recent visuomotor and multimodal robot policies including Diffusion Policy, RT-1, RT-2, CLIPort, PerAct, VIMA, and Octo~\cite{chi2023diffusionpolicy,brohan2023rt1,brohan2023rt2,shridhar2021cliport,shridhar2022peract,jiang2022vima,octo2024octo}. Our formulation is also related to prior work on imitation learning and train--test mismatch, including behavior transformers, DAgger, and scheduled sampling~\cite{shafiullah2022bet,ross2011dagger,bengio2015scheduled}.

\section{Methods}
\label{sec:methods}

\subsection{Problem Formulation}
We consider multi-step manipulation in interactive kitchen environments. At each control step $t$, the robot receives observations $o_t$ and performs a continuous action consisting of an end-effector pose delta and a gripper command. Successful behavior depends on predicting near-future transitions such as contact and articulation. We therefore study whether short-horizon future hypotheses can provide a useful structured prior for decision making and reinforcement-learning-based policy adaptation.

\subsection{Future-Experience Conditioning (FEC)}
\label{sec:fec}

FEC conditions a reactive controller on recent observations and a latent representation of a short-horizon future. During training, the future segment is taken from demonstrations, whereas at test time it is provided by our video generation pipeline.

\paragraph{Per-frame future embeddings}
Let $v^{\mathrm{fut}}_{1:T_f}$ denote the short-horizon future video segment available in a training sample, with $T_f$ future frames. A frame-wise encoder $E(\cdot)$ produces per-frame embeddings:
\begin{equation}
e_{1:T_f} = E\!\left(v^{\mathrm{fut}}_{1:T_f}\right), \qquad e_i \in \mathbb{R}^{d_e}.
\end{equation}

\paragraph{Temporal binning and projection}
We compress the future embeddings $e_{1:T_f}$ into $B$ temporal bins using adaptive average pooling along time:
\begin{equation}
u_{1:B} = \mathrm{Pool}_B\!\left(e_{1:T_f}\right), \qquad u_b \in \mathbb{R}^{d_e}.
\end{equation}
Concatenating the bin vectors and applying a learned projection produces a fixed-dimensional future-conditioning vector:
\begin{equation}
g = P\!\left([u_1;\dots;u_B]\right), \qquad g \in \mathbb{R}^{d_g}.
\end{equation}
This $g$ is used to condition the policy during training and inference. Temporal misalignment is expected to matter because future conditioning is computed from frame-wise embeddings before binning and projection: shifting the future sequence changes which embeddings fall into each temporal bin and therefore changes the resulting future latent.

\paragraph{Temporal-shift protocol}
A natural robustness test in this formulation is to offset the future sequence by a small number of frames \emph{before} temporal binning and projection, then recompute the future-conditioning vector from the shifted frame-wise embeddings. This is the appropriate way to test temporal misalignment in our interface, since perturbing the final pooled latent would mix temporal corruption with representation corruption. Because the controller operates on temporally pooled future embeddings rather than raw frames, mild offsets are expected to be less disruptive than larger ones. In particular, with $T=16$ and $B=4$, a shift of about two frames can still preserve most semantic content within adjacent bins, whereas larger offsets change bin assignments more substantially.

\paragraph{Train and test future source}
Our framework trains the policy with a future experience signal in addition to an observation history, but its source differs between training and inference. During \textbf{inference}, ground-truth future frames are unavailable, so we replace them with \textbf{generated} future clips produced by the DT rollout followed by mask-free diffusion inpainting. Our experiments report controlled conditions that replace the future source at test time (ground-truth vs.\ generated) to quantify the train--test gap.

\subsection{Framework Instantiation of FEC}
Our framework instantiates FEC by constructing a future clip $v_{1:T}$ via an LLM-guided task-grounding step, a robot-free DT articulation rollout, and robot inpainting by a diffusion model.

\subsubsection{Task Grounding and DT Goal Determination}
Given the current scene state and a natural-language command $c$, our framework initializes a task ontology for grounding the manipulation goal. An LLM reasoner operating over this ontology infers a compact structured task specification, including the manipulated object, relevant interaction part, desired state transition, and rollout constraints. This structured output is then deterministically mapped to the DT articulation rollout and does not directly generate low-level robot actions. In all experiments, we use \textbf{GPT-4o} as the back-end reasoner. The scene state used for task grounding is obtained from the simulator at task initialization and is used only for ontology initialization and DT rollout generation. In the current implementation, GPT-4o is used in a structured JSON pipeline rather than free-form text generation. Concretely, it is used to enrich and verify ontology entries from rendered instance images, to fill the task-grounding schema from the command and initialized scene state, and optionally to validate task completion from a screenshot. For planning and verification we use deterministic decoding, while ontology enrichment uses a small nonzero temperature. When verification fails to parse, the system falls back to the sanitized ontology produced by the rule-based lint pass.

\subsubsection{Robot-Free Digital Twin Articulation Video}
From the ontology-grounded task specification corresponding to command $c$, a DT rollout generates a transparent video $\tilde{v}_{1:T}$ showing only the environment and target object motion from target camera viewpoints. This specifies the intended object evolution without requiring the DT to synthesize robot kinematics or contacts.

\subsubsection{Robot Inpainting with Video Diffusion}
Given $\tilde{v}_{1:T}$, a video diffusion inpainting model generates a robot-containing future clip $v_{1:T}$ that is visually consistent with the DT dynamics and camera angle. Our inpainting model is built on the CogVideoX diffusion backbone~\cite{yang2024cogvideox}, using the VideoPainter framework~\cite{bian2025videopainter} as a starting point. We modify the pipeline to be mask-free and robot-conditioned: the model receives only an initial image containing the robot and the DT ``transparent'' video, and outputs a robot-inpainted future clip. Training uses paired clips, with robot-free DT video as conditioning and the corresponding robot-execution clip as target reconstruction. We condition on the first frame to anchor robot appearance and initial pose.

\begin{figure*}[t]
  \centering
  \includegraphics[width=\textwidth]{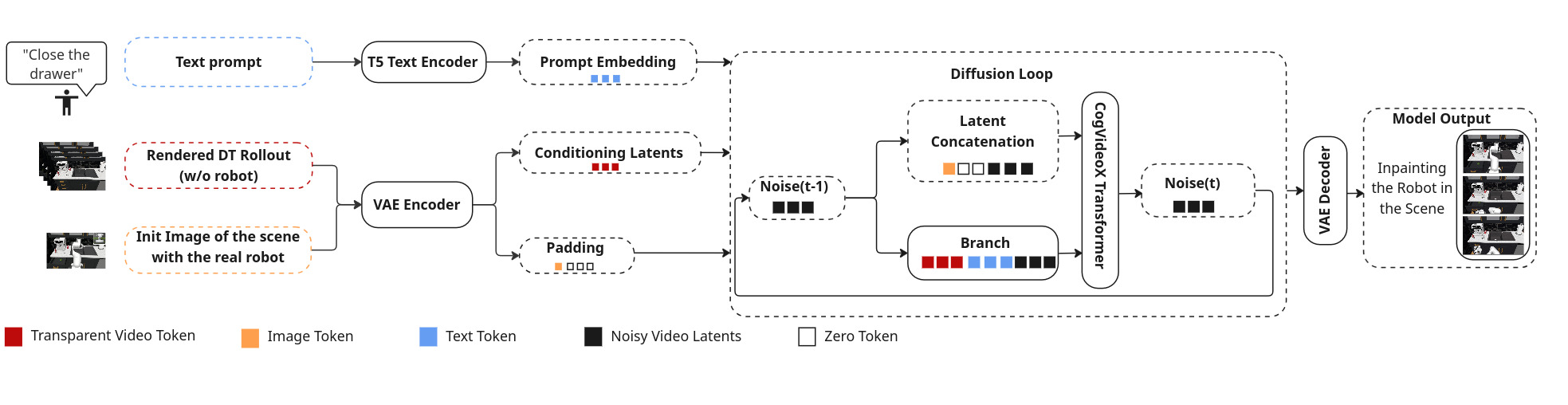}
  \caption{Mask-free diffusion model used for robot inpainting. Conditioning inputs include an initial image with the robot, a text prompt, and the DT transparent video. The model uses a CogVideoX backbone~\cite{yang2024cogvideox} with a VideoPainter-style conditioning branch~\cite{bian2025videopainter}.}
  \label{fig:dm_arch}
\end{figure*}

\paragraph{Mask-free 32-channel branch}
To avoid requiring privileged robot masks at inference, we train a mask-free variant based on CogVideoX~\cite{yang2024cogvideox} and our adaptation of VideoPainter~\cite{bian2025videopainter}. The inpainting branch uses a 32-channel configuration. Thus, \emph{no robot masks or segmentations are provided at inference}, and the model must infer robot pixels and motion from DT dynamics plus the initial frame anchor. The architecture of this mask-free diffusion model is shown in Fig.~\ref{fig:dm_arch}.

\subsection{BC and BC with RL Fine-Tuning}
\label{sec:bcrl_policy}

Our main controller instantiation is a behavior-cloning pipeline followed by reinforcement-learning fine-tuning. The BC policy maps the recent observation history and future-conditioning vector to an action,
\begin{equation}
a_t^{\mathrm{BC}} = \pi_{\phi}^{\mathrm{BC}}(o_{t-K:t}, g_t, c),
\end{equation}
which defines the \textbf{BC-only} baseline.

We then initialize the policy from the BC solution and further optimize it with reinforcement learning:
\begin{equation}
a_t = a_t^{\mathrm{BC}} + \Delta a_t,
\end{equation}
where $\Delta a_t$ denotes the policy change induced by RL fine-tuning from the BC initialization. When future conditioning is enabled, the RL-fine-tuned policy uses the same future-conditioning vector $g_t$ as the BC initialization.

\subsection{Future-Conditioned Streaming Flow Policy Baseline}
We also adapt a Streaming Flow Policy (SFP)~\cite{jiang2025sfp} as a comparison baseline, conditioning it on both the recent past and the same future signal used by our BC and BC+RL controllers. At time $t$, the policy conditions on a short observation history $o_{t-K:t}$ and a future-conditioning vector $g$:
\begin{equation}
a_t \sim \pi_{\theta}\!\left(\cdot \mid o_{t-K:t},\, g,\, c\right).
\end{equation}

\subsection{Training Objective}
We train SFP using a conditional flow-matching objective over short action chunks. For a sampled action chunk $\sigma \in \mathbb{R}^{T_{\mathrm{train}}\times d_a}$ and its finite-difference velocity $\dot{\sigma}$, we sample a continuous time $\lambda \sim \mathcal{U}[0,1]$ and noise $\epsilon \sim \mathcal{N}(0,I)$, and define:
\begin{equation}
\eta(\lambda)=\sigma_0 e^{-k\lambda}\epsilon,\qquad
a = \sigma + \eta(\lambda),\qquad
v = \dot{\sigma} - k\,\eta(\lambda).
\end{equation}
The network predicts a velocity field $v_\theta(a,\lambda\mid o_{t-K:t}, g)$ and is trained with an MSE loss:
\begin{equation}
\min_{\theta}\;
\mathbb{E}\Big[
\big\|v_\theta(a,\lambda\mid o_{t-K:t}, g)-v\big\|_2^2
\Big].
\end{equation}

\subsection{Training Procedure}
We train the VDM on paired DT/robot video clips using fixed-length windows and first-frame anchoring. Our main controller pipeline is based on behavior cloning followed by RL fine-tuning. We first train a BC-only controller by supervised imitation on demonstrations, then continue optimization with RL fine-tuning in the simulator to obtain BC+RL. At inference time, we replace demonstration futures with generated futures from our framework to match deployment. For comparison, we also train an SFP baseline using the same future-conditioning interface.

\section{Experiments}
\label{sec:experiments}

\subsection{Experimental Setup}
\label{sec:exp_setup}

\begin{figure*}[!t]
  \centering
  \includegraphics[width=\textwidth]{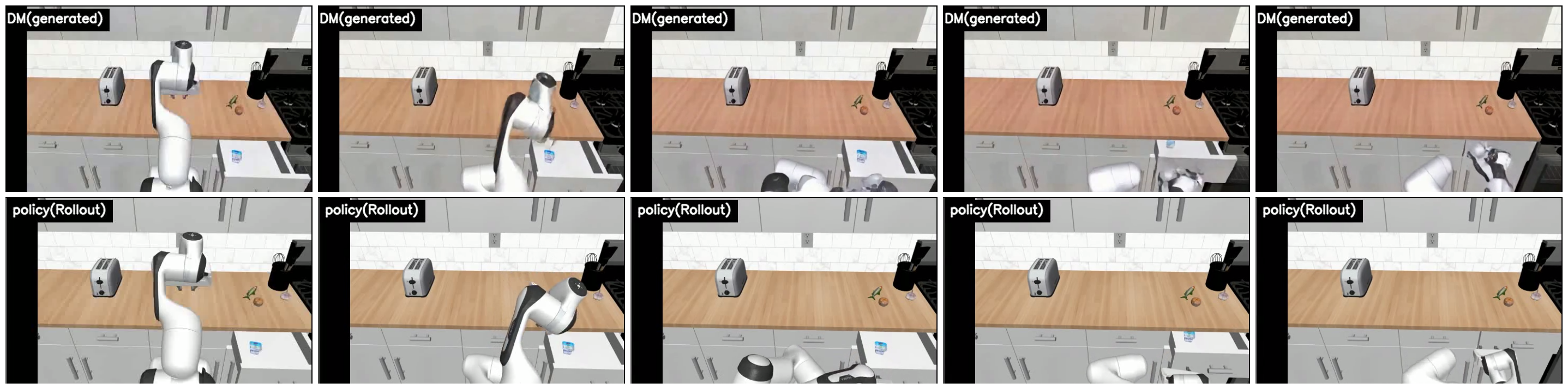}
  \caption{Qualitative example on RoboCasa. Top: future sequence generated by the video diffusion model (VDM). Bottom: rollout of the robot policy conditioned on that generated future.}
  \label{fig:qual_robocasa}
\end{figure*}

\begin{figure*}[!t]
  \centering
  \includegraphics[width=\textwidth]{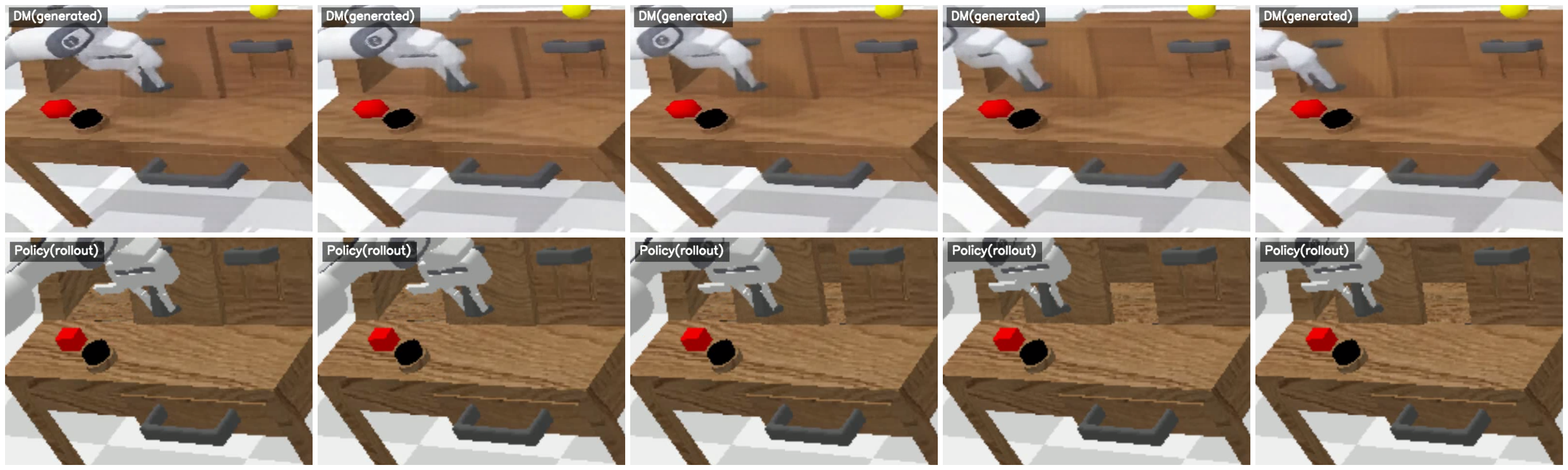}
  \caption{Qualitative examples on CALVIN. Top: future sequence generated by the video diffusion model (VDM). Bottom: rollout of the robot policy conditioned on that generated future.}
  \label{fig:qual_calvin}
\end{figure*}

We emphasize that these experiments are a simulation study of the future-conditioning interface and its effect on policy execution and RL fine-tuning; they are not intended as a full evaluation of digital-twin fidelity, raw-perception scene understanding, or sim-to-real transfer.

\noindent\textbf{Benchmarks and tasks.}
We evaluate our framework on \textbf{RoboCasa} and \textbf{CALVIN}. On \textbf{RoboCasa}, we evaluate \textbf{CloseDrawer}. On \textbf{CALVIN}, we evaluate \texttt{open\_drawer}, \texttt{close\_drawer}, \texttt{turn\_on\_lightbulb}, \texttt{turn\_off\_lightbulb}, \texttt{turn\_on\_led}, \texttt{turn\_off\_led}, \texttt{push\_into\_drawer}, and \texttt{move\_slider\_left}.

\noindent\textbf{Exploration viewpoint.}
In BC+RL, the four future conditions also define four exploration regimes. \textbf{NoFuture} removes explicit future guidance, \textbf{GTFuture} provides an oracle short-horizon prior, \textbf{GenFuture} provides the realistic imperfect prior available at deployment, and \textbf{WrongFuture} provides an intentionally misleading prior. This lets us test whether future clips help RL fine-tuning because they provide structured exploration guidance rather than because they simply add extra input.

\noindent\textbf{Scope of grounding.}
The task-grounding module is used only once at task initialization to create the ontology-grounded DT rollout. During closed-loop execution, the controller does not repeatedly query the LLM reasoner or update the ontology online; it acts from recent observations together with the future-conditioning vector.

\noindent\textbf{Implementation details.}
In all experiments, the future clip contains $T=16$ frames. We compress per-frame future embeddings into $B=4$ temporal bins and project them to a latent of dimension $d_g=256$. For the BC/BC+RL controllers, future frames are encoded frame-wise with a ResNet-18 backbone pretrained on ImageNet before temporal binning and projection. Our main RL experiments use BC-initialized TD3-style fine-tuning with batch size 256, discount 0.99, target update $\tau=0.005$, policy delay 2, actor and critic learning rates $5\times10^{-5}$, policy noise 0.05, noise clip 0.08, and 140k environment steps. For task grounding, GPT-4o is used once at task initialization to fill the structured ontology-based schema that specifies the manipulated object, interaction part, and desired state transition for DT rollout generation. The LLM reasoner is queried through a structured JSON interface rather than inside the per-step control loop; planning and verification use deterministic decoding, while ontology enrichment uses a small nonzero temperature. To characterize the compute cost of the future-generation component, our mask-free video diffusion model was trained on a single NVIDIA A40 GPU with peak training memory of 44\,GB; at inference, generating one 16-frame future clip with 20 denoising steps takes approximately 40--45\,s in bf16 precision on the same hardware. These measurements indicate that the future-generation module operates in a practical single-GPU setting, although we do not claim a full efficiency benchmark. The LLM-based reasoner is outside the closed-loop policy execution path.

\noindent\textbf{Observations and control.}
Policies operate from RGB observations and output continuous end-effector actions with a gripper command. At each control step $t$, the controller receives the observation history $o_{t-K:t}$ and the future-conditioning vector $g_t$, and outputs an action $a_t \in \mathbb{R}^{7}$ consisting of a 6-DoF end-effector pose delta and a gripper command.

\noindent\textbf{Future sources and training.}
We compare ground-truth futures extracted from demonstrations and generated futures produced by our framework. Generated futures are constructed in two stages: a robot-free DT rollout specifies the intended object motion, and a mask-free video diffusion inpainting model generates a robot-containing future clip consistent with that rollout. We train the VDM on paired DT/robot clips using fixed-length clips and first-frame anchoring. We then evaluate a BC-only controller and a BC+RL controller, with a future-conditioned SFP baseline for comparison. For each controller, we train three independent models with random seeds 42, 43, and 44.

\noindent\textbf{Metrics.}
We report task success rate (\%) under NoFuture, GTFuture, GenFuture, and WrongFuture. Each evaluation episode runs for at most 200 control steps.

\subsection{Baselines and Ablations}
\label{sec:exp_ablations}

We evaluate the following baselines and ablations:
\begin{itemize}
  \item \textbf{Controllers:} BC-only, BC+RL, and a future-conditioned SFP comparison baseline.
  \item \textbf{NoFuture:} no future-conditioning signal.
  \item \textbf{GTFuture:} ground-truth future clips at test time.
  \item \textbf{GenFuture:} generated futures from our framework at test time.
  \item \textbf{WrongFuture:} intentionally mismatched futures to verify that gains are not due simply to extra input.
\end{itemize}

\subsection{Results}
\label{sec:exp_results}

\begin{table*}[!t]
  \centering
  \scriptsize
  \setlength{\tabcolsep}{3.2pt}
  \renewcommand{\arraystretch}{1.02}

  \begin{minipage}[t]{0.49\textwidth}
    \centering
    \textbf{(a) CALVIN: SFP}
    \vspace{1mm}

    \begin{tabular}{lcccc}
      \toprule
      \textbf{Result} & \textbf{NoF} & \textbf{GT} & \textbf{Gen} & \textbf{Wrong} \\
      \midrule
      8-task average & $6.1 \pm 4.1$ & $24.3 \pm 4.7$ & $17.8 \pm 2.9$ & $0.0 \pm 0.0$ \\
      \bottomrule
    \end{tabular}
  \end{minipage}\hfill
  \begin{minipage}[t]{0.49\textwidth}
    \centering
    \textbf{(b) CALVIN: BC+RL}
    \vspace{1mm}

    \begin{tabular}{lcccc}
      \toprule
      \textbf{Result} & \textbf{NoF} & \textbf{GT} & \textbf{Gen} & \textbf{Wrong} \\
      \midrule
      8-task average & $54.0 \pm 3.2$ & $89.1 \pm 1.6$ & $69.8 \pm 2.1$ & $0.0 \pm 0.0$ \\
      \bottomrule
    \end{tabular}
  \end{minipage}

  \vspace{2mm}

  \begin{minipage}[t]{0.49\textwidth}
    \centering
    \textbf{(c) CALVIN: BC-only}
    \vspace{1mm}

    \begin{tabular}{lcccc}
      \toprule
      \textbf{Result} & \textbf{NoF} & \textbf{GT} & \textbf{Gen} & \textbf{Wrong} \\
      \midrule
      8-task average & $39.8 \pm 4.3$ & $61.1 \pm 0.8$ & $53.6 \pm 1.9$ & $0.0 \pm 0.0$ \\
      \bottomrule
    \end{tabular}
  \end{minipage}\hfill
  \begin{minipage}[t]{0.49\textwidth}
    \centering
    \textbf{(d) RoboCasa: CloseDrawer}
    \vspace{1mm}

    \begin{tabular}{lcccc}
      \toprule
      \textbf{Method} & \textbf{NoF} & \textbf{GT} & \textbf{Gen} & \textbf{Wrong} \\
      \midrule
      SFP     & $36.7 \pm 2.5$ & $58.3 \pm 2.1$ & $49.7 \pm 2.3$ & $0.0 \pm 0.0$ \\
      BC-only & $52.3 \pm 2.1$ & $74.7 \pm 2.5$ & $66.3 \pm 2.1$ & $0.0 \pm 0.0$ \\
      BC+RL   & $61.7 \pm 2.5$ & $82.3 \pm 2.1$ & $75.7 \pm 2.3$ & $0.0 \pm 0.0$ \\
      \bottomrule
    \end{tabular}
  \end{minipage}

  \par\vspace{2mm}
  \begin{minipage}[t]{0.49\textwidth}
    \centering
    \textbf{(e) RoboCasa: OpenSingleDoor}
    \vspace{1mm}

    \begin{tabular}{lcccc}
      \toprule
      \textbf{Method} & \textbf{NoF} & \textbf{GT} & \textbf{Gen} & \textbf{Wrong} \\
      \midrule
      SFP     & $0.0 \pm 0.0$ & $5.1 \pm 1.2$ & $0.0 \pm 0.0$ & $0.0 \pm 0.0$ \\
      BC-only & $33.1 \pm 2.8$ & $63.2 \pm 1.9$ & $56.3 \pm 3.3$ & $0.0 \pm 0.0$ \\
      BC+RL   & $46.1 \pm 1.3$ & $73.1 \pm 1.7$ & $66.2 \pm 1.2$ & $0.0 \pm 0.0$ \\
      \bottomrule
    \end{tabular}
  \end{minipage}

  \caption{Success rate (\%) under four future conditions: NoFuture (NoF), GTFuture (GT), GenFuture (Gen), and WrongFuture (Wrong). Results are averaged over 3 training seeds.}
  \label{tab:all_results}
\end{table*}

\noindent\textbf{Summary of quantitative results.}
Across both benchmarks, our main BC+RL instantiation is strongest overall. The future-conditioned SFP baseline benefits from the same interface but remains below BC+RL. Generated future conditioning improves over NoFuture, whereas WrongFuture degrades sharply. This suggests that task-consistent future conditioning can support policy execution and reinforcement-learning-based adaptation, whereas mismatched future signals can be harmful.

\begin{figure}[!t]
  \centering
  \includegraphics[width=0.94\columnwidth]{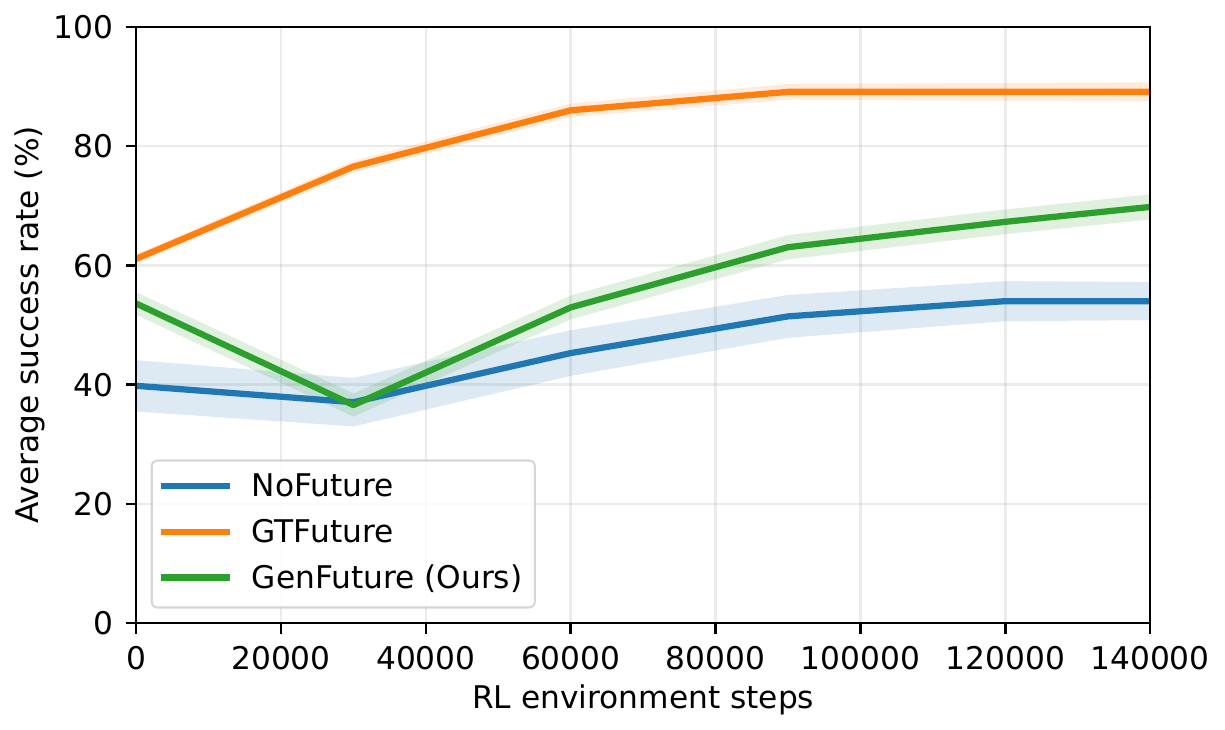}
  \caption{Average BC+RL learning curves across 8 CALVIN tasks under four future conditions. Curves show mean success rate over RL environment steps, averaged across tasks, and shaded regions indicate standard deviation over 3 seeds. GTFuture improves fastest and reaches the highest average performance, GenFuture also improves earlier and to a higher level than NoFuture, while WrongFuture remains at zero throughout training. Final success is 89.1\% for GTFuture, 69.8\% for GenFuture, and 54.0\% for NoFuture, consistent with Table~\ref{tab:all_results}.}
  \label{fig:rl_exploration_curve}
  \vspace{-2mm}
\end{figure}

\noindent\textbf{Exploration efficiency during RL fine-tuning.}
To test whether generated future conditioning improves RL adaptation beyond final execution alone, we analyze BC+RL learning curves averaged across 8 CALVIN tasks over the full 140k RL environment steps. Figure~\ref{fig:rl_exploration_curve} shows that GTFuture improves fastest and reaches the highest overall performance, GenFuture also improves earlier and to a higher level than NoFuture, and WrongFuture remains at zero throughout training. Final success is 89.1\% for GTFuture, 69.8\% for GenFuture, and 54.0\% for NoFuture. This provides broader evidence, beyond a single task, that informative future conditioning can improve RL fine-tuning efficiency, while mismatched future signals can strongly hinder adaptation.

\noindent\textbf{Interpretation as structured exploration.}
Accurate futures accelerate adaptation most, realistic but imperfect futures remain useful, the absence of future guidance slows learning, and mismatched futures can fully misguide behavior.

\noindent\textbf{Limitations and failure modes.}
Our study is limited to simulation and does not evaluate sim-to-real transfer or real-robot performance. The task-grounding stage uses privileged simulator state at task initialization rather than raw sensory reconstruction alone, and we do not provide a dedicated quantitative study of digital-twin or video-generation fidelity. Failures usually arise when the generated future contains inaccurate contact geometry or misleading interaction cues.

\section{Conclusion}
\label{sec:conclusion}

We presented a simulation study of future-conditioned control for multi-step robot manipulation. Our main controller instantiation is a behavior-cloning pipeline with reinforcement-learning fine-tuning, while a future-conditioned SFP model is included as a comparison baseline. Our framework combines LLM-guided task grounding, a robot-free digital-twin rollout, and mask-free diffusion-based robot inpainting to generate short-horizon future clips that are used as conditioning signals for control. Across RoboCasa and CALVIN, task-consistent future conditioning improves performance over no-future conditioning in our experiments, while mismatched futures are harmful, and our BC+RL instantiation achieves the strongest overall results. In addition, an average BC+RL learning-curve analysis across 8 CALVIN tasks shows that GTFuture improves fastest, GenFuture also improves earlier and to a higher level than NoFuture, and WrongFuture remains at zero throughout training.

These results support the view that short-horizon future videos can serve as useful structured priors for policy execution and RL adaptation under imperfect future predictions. 

\section*{ACKNOWLEDGMENT}
The authors thank their collaborators and reviewers for helpful feedback.

\bibliographystyle{IEEEtran}
\bibliography{references}

\end{document}